\documentclass[letterpaper, 10 pt, conference]{ieeeconf}

\usepackage{amsmath}
\usepackage{amssymb}

\usepackage{tabularx}
\usepackage{graphicx}
\usepackage{bm}
\usepackage{color}
\usepackage{float}
\usepackage{units}
\usepackage{url}
\usepackage{hyperref}
\usepackage{balance}
\usepackage{amsmath}

\makeatletter 
\def\endfigure{\end@float} 
\def\endtable{\end@float}
\makeatother

\usepackage{subcaption}
\usepackage{caption}

\renewcommand{\unit}[1]{{\rm #1} }

\newcommand{\update}[1]{\textcolor{black}{#1}}

\IEEEoverridecommandlockouts

\begin{document} 

\title{\Large \bf 			
Dynamic Walking of Bipedal Robots on Uneven Stepping Stones via Adaptive-frequency MPC
}

\author{Junheng Li and Quan Nguyen\thanks{Junheng Li and Quan Nguyen are with the Department of Aerospace and Mechanical Engineering, University of Southern California, Los Angeles, CA 90089.
email:{\tt\small junhengl@usc.edu, quann@usc.edu}}%
}%
	
\maketitle

\begin{abstract}

This paper presents a novel Adaptive-frequency MPC framework for bipedal locomotion over terrain with uneven stepping stones. 
In detail, we intend to achieve adaptive foot placement and gait period for bipedal periodic walking gait with this MPC,in order to traverse terrain with discontinuities without slowing down. We pair this adaptive-frequency MPC with a kino-dynamics trajectory optimization for optimal gait periods, center of mass (CoM) trajectory, and foot placements. 
We use whole-body control (WBC) along with adaptive-frequency MPC to track the optimal trajectories from the offline optimization.
In numerical validations, our adaptive-frequency MPC framework with optimization has shown advantages over fixed-frequency MPC.
The proposed framework can control the bipedal robot to traverse through uneven stepping stone terrains with perturbed stone heights, widths, and surface shapes while maintaining an average speed of 1.5 $\unit{m/s}$.

\end{abstract}


\section{Introduction}
\label{sec:Introduction}

Uneven terrain locomotion has always been one of the most important problems that researchers aim to solve on  bipedal robots via motion planning and control. The value of such capability will allow bipedal robots to perform robust locomotion in many real-world tasks such as rescue and exploration missions with unknown terrains. Recent advancement in control strategies has allowed many successful integrations of control frameworks with bipedal robots. 

For instance, on one hand, Hybrid Zero Dynamics (HZD) model \cite{westervelt2003hybrid} is an effective control scheme employed on bipedal robots such as MABEL \cite{sreenath2011compliant}. 
HZD on ATRIAS robot \cite{rezazadeh2015spring} has allowed more intricate motion planning strategies to be integrated, such as gait libraries for stepping stones \cite{nguyen2018dynamic}. The gait library collected from offline optimization has allowed ATRIAS (2-D) to precisely place its foot on the stepping stones by online motion planning and position control. This position-control-based approach requires accurate terrain information, including next stone distance and height, and is not robust to uneven terrain perturbations. 

On the other hand, force-based control schemes on quadruped robots became more popular. Such control frameworks can be used with linearized dynamics models and constraints. The Quadratic Programming (QP)-based force control and Model Predictive Control (MPC)  on quadruped robots (\cite{di2018dynamic,nguyen2019optimized}) both employ simplified rigid-body dynamics and have demonstrated effectiveness in stable locomotion over uneven terrain.
We believe bipedal robots can also benefit from the robustness on uneven terrain with force-based locomotion control. 

\begin{figure}[t]
		\center
		\includegraphics[width=1 \columnwidth]{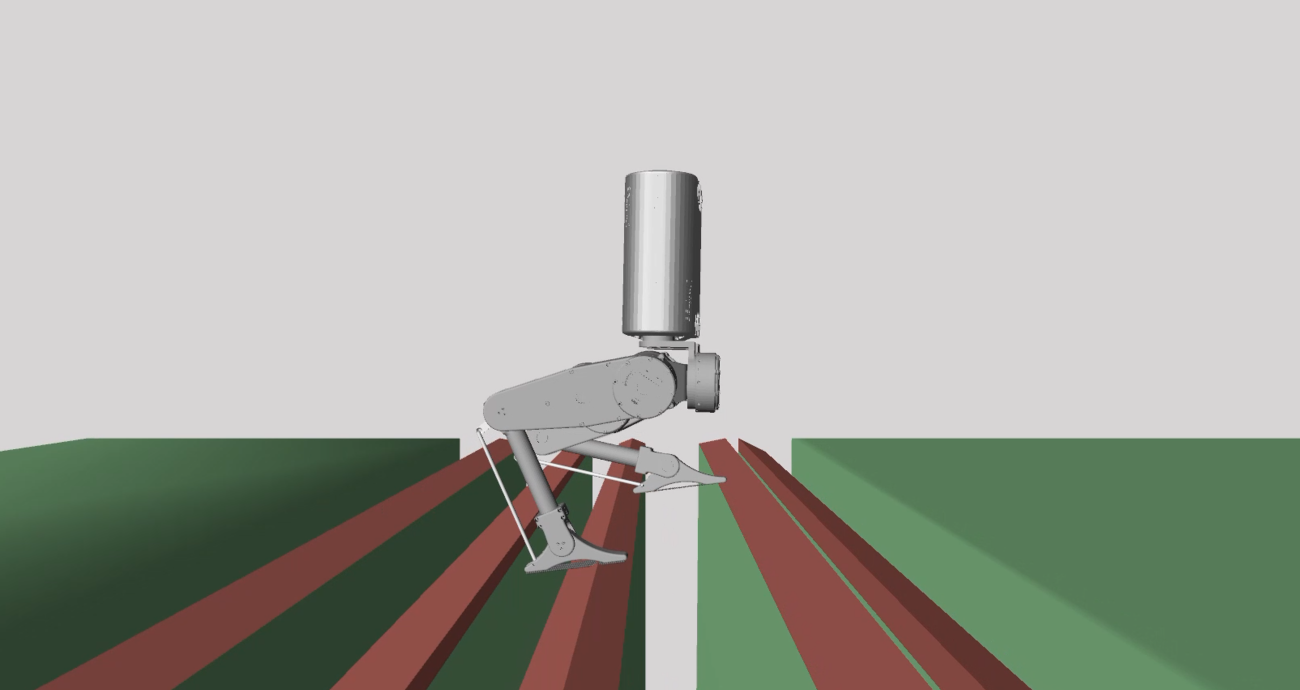}
		\caption{{\bfseries Bipedal Robot Traversing Terrain with Uneven Stepping Stones}  Simulation video: \protect\url{https://youtu.be/8hLihy96lCg}. }
		\label{fig:title}
		\vspace{-1.5em}
\end{figure}

Our recent work on force-and-moment-based MPC schemes on a 16-Degree-of-Freedom (DOF) bipedal robot \cite{li2021force} has allowed stable 3-D locomotion with fixed gait periods (i.e. fixed-frequency MPC). \update{However, due to the unawareness of the terrain, the robot cannot adapt its footsteps based on the terrain.} The next-step foot placement \cite{raibert1986legged} of bipedal locomotion is dependent on both linear velocity and gait period. Hence, when maintaining a constant velocity during walking while aiming to vary step length, it can be achieved by adjusting the gait period of each step. We introduce adaptive frequency to the MPC to allow the robot to walk with varied gait periods for each step and achieve varied step lengths with a constant walking speed. 

 Kino-dynamics-based trajectory optimization has been introduced and used in many works on mobile-legged robots (e.g. \cite{dai2014whole, herzog2016structured}). The framework has the advantage of simplified system dynamics while being able to apply robot joint constraints. To synchronize the motion control and optimization, we use the same simplified dynamics model in both MPC and optimization, the same foot placement policy in swing foot control and optimization foot placement, and the same discrete time steps in MPC and optimization.
 
 Many related works (e.g.,\cite{kryczka2015online,khadiv2016step,guo2021fast,daneshmand2021variable}) that use trajectory optimization/planning for bipedal gait and trajectory generation share a similarity in that the foot placement adaptation is included in the frameworks to optimize best capture point locations. In our work, to allow bipedal robots to overcome very narrow stepping stones, exact foot placement on the stone is required. We pre-define the desired step locations in optimization to optimize the gait periods and CoM trajectory based on each stride length. 

Tracking optimal trajectory with only MPC is not optimal due to its inherent low sampling frequency, which is even lower with a long gait period. We pair the MPC with a higher-frequency Whole-body Control (WBC) for more accurate trajectory tracking. 
MIT Mini Cheetah \cite{katz2019mini,kim2019highly} quadruped robot and MIT Humanoid robot \cite{chignoli2021humanoid} both have demonstrated outstanding balancing performance during dynamical motion with the force-based MPC and WBC combination. We develop the WBC strategy to work with our bipedal force-and-moment-based MPC. WBC in \cite{kim2020dynamic} employed on bipedal robots \cite{kim2016stabilizing, ahn2019control} validated the feasibility of a WBC-type control strategy in dynamic locomotion with periodic gaits. In our approach, We combine Kino-dynamics trajectory optimization with adaptive-frequency MPC framework for bipedal robot traversing stepping stones and use WBC as low-level force-to-torque mapping and trajectory tracking control.

The main contributions of the paper are as follows:
\begin{itemize}
    \item \update{We allow the bipedal to have adaptive foot placement and gait periods for each step, and realize it in control with adaptive-frequency MPC as our main locomotion controller.}
    
    \item \update{We enhance the adaptive-frequency MPC by kino-dynamics trajectory optimization for optimal trajectory generation and WBC as tracking control.}
    
    \item \update{We use the proposed framework in bipedal locomotion over uneven stepping stones. The proposed method allows the bipedal robot to maintain high speed at around 1.5 $\unit{m/s}$ when traversing uneven stepping stone terrains with height, width, and stone surface shape perturbations while only requiring minimal terrain knowledge.}
\end{itemize}

The rest of the paper is organized as follows. Section. \ref{sec:robotModel} introduces the physical design parameters of the bipedal robotand the overview of the system architecture including optimization and control. Section.\ref{sec:trackingControl} presents the adaptive-frequency trajectory optimization framework with the bipedal kino-dynamics model. Section. \ref{sec:trackingControl} presents the adaptive-frequency MPC framework. Some simulation result highlights and comparisons are presented in Section. \ref{sec:Results}.


\section{Bipedal Robot Model and System Overview}
\label{sec:robotModel}

\subsection{Bipedal Robot Model}

\begin{figure}[!h]
\vspace{0.1cm}
		\center
		\includegraphics[width=1 \columnwidth]{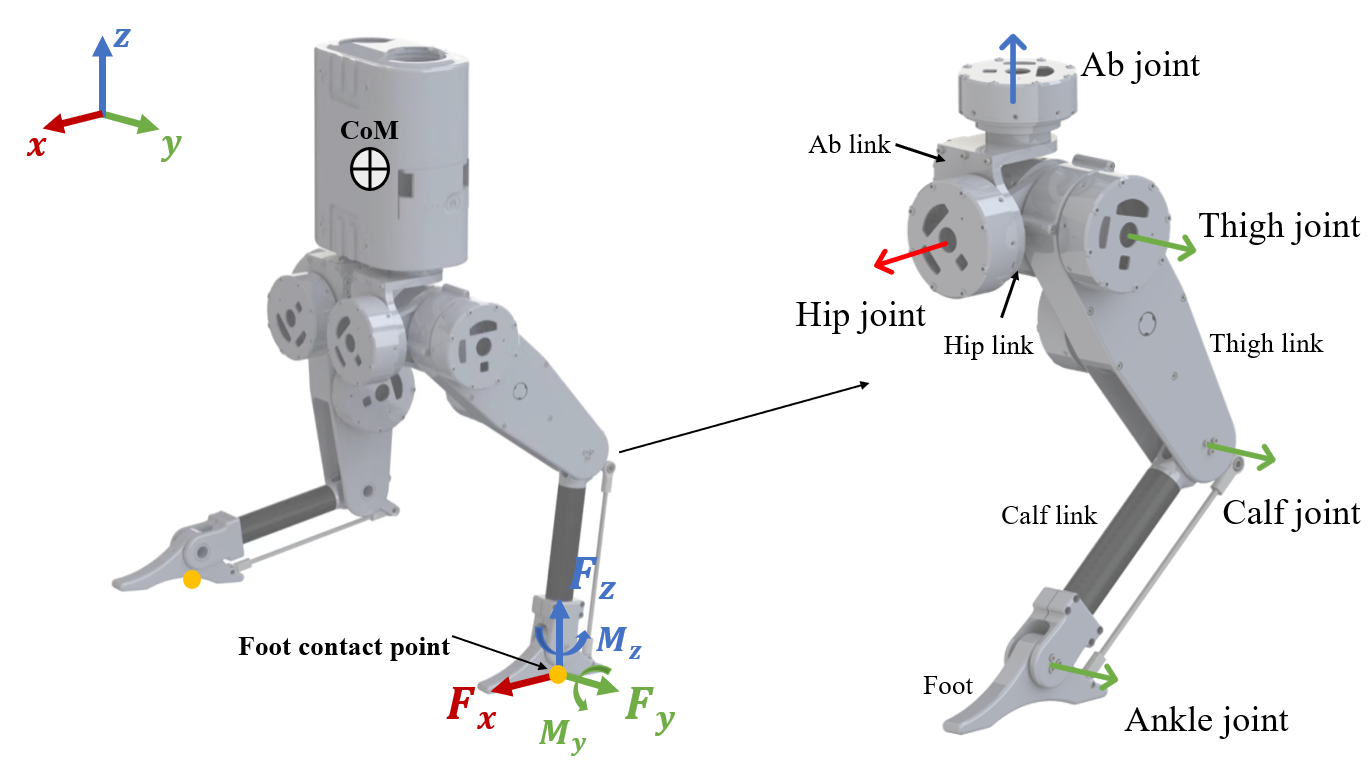}
		\caption{{\bfseries Bipedal Robot Configuration and Simplified Dynamics Model}}
		\label{fig:design}
		\vspace{-0.5em}
\end{figure}

In this section, we present the bipedal robot model that is used for this work. Our bipedal robot model is enhanced from our previous design in \cite{li2021force}, a small-scale bipedal robot with 5-DoF legs. Presented in Figure. \ref{fig:design}, each of the robot legs consists of ab/ad, hip, thigh, calf, and ankle joints which are all actuated by Unitree A1 torque-controlled motor. A1 motor is a powerful joint motor with a 33.5 $\unit{Nm}$ maximum torque output and 21.0 $\unit{rad/s}$ maximum joint speed output while weighing only 0.6 $\unit{kg}$.

In this bipedal leg design, we strategically placed all joint actuators on the upper of the thigh links, close to the hips, for mass concentration, in order to minimize the leg dynamics during locomotion. Negligible leg mass is an important assumption in our force-and-moment-based simplified dynamics model in MPC \cite{li2021force}. The trunk mass of the bipedal robot is 5.8 $\unit{kg}$ and the overall mass is around 11 $\unit{kg}$. 
More details about the physical design parameters can also be found in \cite{li2021force}.

\subsection{System Overview}
\label{sec:sysoverview}

\begin{figure}[!h]
	\vspace{-0.3cm}
		\center
		\includegraphics[width=1 \columnwidth]{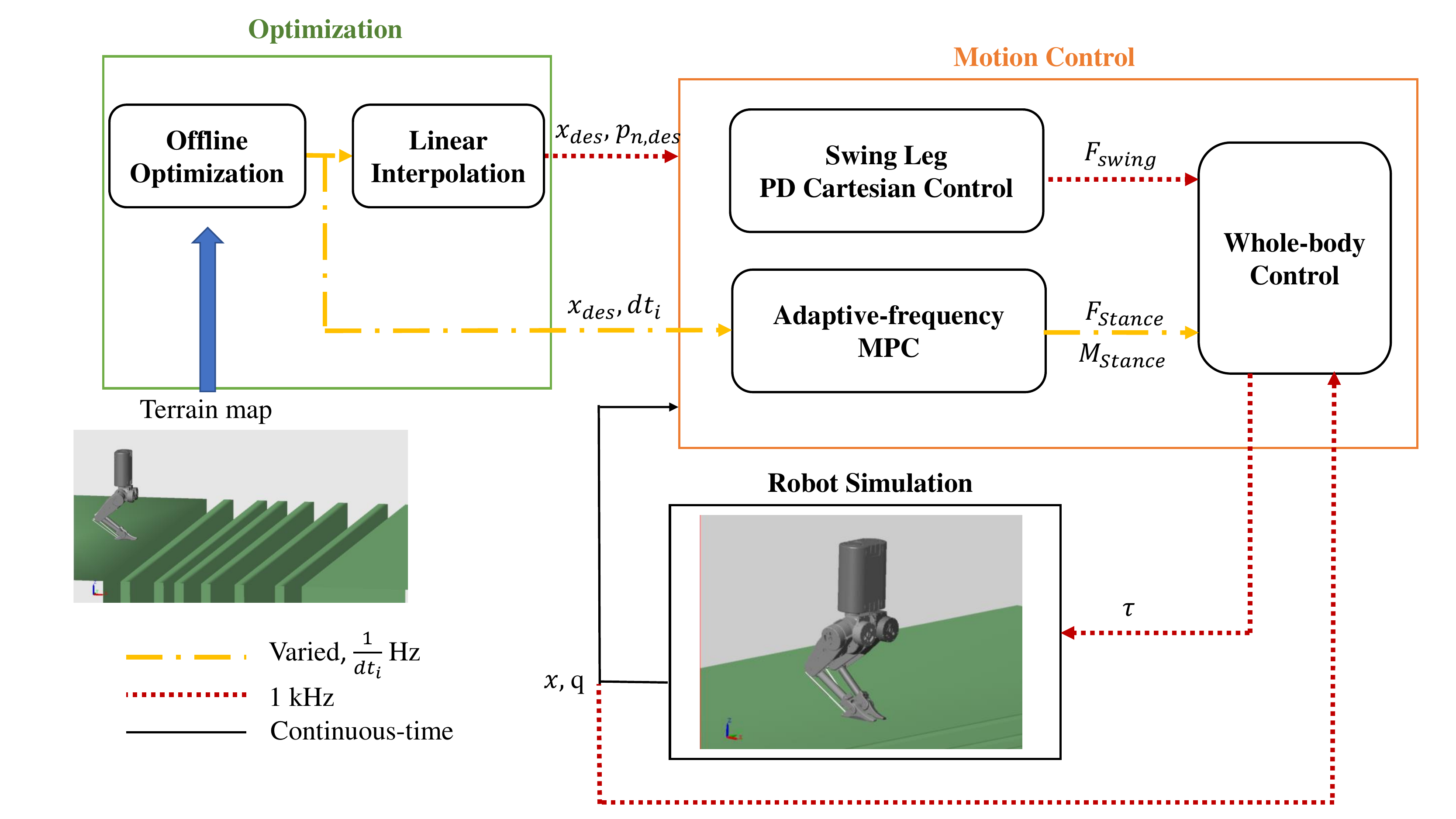}
		\caption{{\bfseries System Block Diagram} Optimization and control architecture.}
		\label{fig:controlArchi}
		\vspace{-.2cm}
\end{figure}

The optimization and control system block diagram is shown in Figure. \ref{fig:controlArchi}. 
We aim to achieve varied step lengths for each step in bipedal locomotion by varying gait frequencies in adaptive-frequency MPC. The proposed framework is built around this controller. In order to allow more stable and efficient locomotion, we pair the MPC control framework with offline trajectory optimization to generate desired trajectories.

The optimization framework uses terrain map to generate discrete optimization data including desired body CoM trajectory $\bm x_{des} \in \mathbb{R}^3$, desired foot position $\bm p_{n,des} \in \mathbb{R}^3$ for $n$th foot, and discrete sampling time $dt_i$ at time step $i$ for MPC. The CoM trajectory and foot positions are linearly-interpolated to have a sampling frequency at 1 $\unit{kHz}$ to match the frequency of swing leg control and WBC. The MPC accepts the optimization data at its native frequency due to the synchronization of sampling time. 
Reaction forces from MPC and swing leg control are input into WBC to be mapped to joint torques $\bm \tau  \in \mathbb{R}^{10}$.
The robot state feedback $\bm x \in \mathbb{R}^{12}$ include body Euler angles (roll, pitch, and yaw) ${\Theta = [\phi,\:\theta,\:\psi]}^\intercal$, position $\bm p_c$, velocity of body CoM  $\dot{\bm p}_c$, and angular velocity $\bm \omega$. Joint feedback $\mathbf q \in \mathbb{R}^{10}$ includes the joint positions of the bipedal robot.

\section{Kino-dynamics-based Adaptive-frequency Trajectory Optimization}
\label{subsec:Optimization}

Humans can walk with different step lengths every step to adapt to the terrain and can allow swing foot to remain in the air for different periods of time. We intend to use this adaptive-frequency trajectory optimization framework to allow bipedal robots to walk with such characteristics. 

We choose the kino-dynamics model in our optimization framework in order to reduce the computation cost compared to using a full-dynamics model. 
The average solving time of offline trajectory optimization in our approach is shown in Table. \ref{tab:solvingTime}.


\subsection{Simplified Dynamics Model}
We first present the force-and-moment-based simplified dynamics model we use in both the kino-dynamics trajectory optimization and adaptive-frequency MPC framework, introduced in the author's previous work \cite{li2021force}. The simplified force-based dynamics model with ground reaction force and moment control inputs is shown in Figure. \ref{fig:design}. The control input consists of
$ \bm u=[\bm F_1;\:\bm F_2;\:\bm M_1;\:\bm M_2]^\intercal \in \mathbb{R}^{10}$, where $ \bm F_n = [ F_{nx},\: F_{ny},\: F_{nz}]^\intercal, \bm M_n = [ M_{ny},\: M_{nz}]^\intercal,$ leg $n = 1, 2 $.

We choose the state variables as $[{\bm \Theta};{\bm p}_c;{\bm \omega};\dot {{\bm p}}_c]$ and control inputs as $\bm u$, then the simplified dynamics equation can be represented as
\begin{align}
\label{eq:simpDyn}
\frac{d}{dt}\left[\begin{array}{c} {\bm \Theta}\\{\bm p}_c\\{\bm \omega}\\\dot {{\bm p}}_c \end{array} \right] 
= \bm A \left[\begin{array}{c} {\bm \Theta}\\{\bm p}_c\\{\bm \omega}\\\dot {{\bm p}}_c \end{array} \right] + \bm B \bm u + \left[\begin{array}{c} \mathbf 0_{3\times1}\\\mathbf 0_{3\times1}\\\mathbf 0_{3\times1}\\\bm g \end{array} \right] 
\end{align}

\begin{align}
\label{eq:A}
\bm A = \left[\begin{array}{cccc} 
\mathbf 0_3 & \mathbf 0_3 & \mathbf R_z & \mathbf 0_3 \\
\mathbf 0_3 & \mathbf 0_3 & \mathbf 0_3 & \mathbf I_3\\
\mathbf 0_3 & \mathbf 0_3 & \mathbf 0_3 & \mathbf 0_3\\
\mathbf 0_3 & \mathbf 0_3 & \mathbf 0_3 & \mathbf 0_3 \end{array} \right], 
\mathbf R_z = \left[\begin{array}{ccc}
{c_\psi} & -{s_\psi} & 0 \\
{s_\psi} & c_\psi & 0 \\
0 & 0 & 1  \end{array} \right] 
\end{align}

\begin{align}
\label{eq:B}
\bm B = \left[\begin{array}{ccccc} 
\mathbf 0_3 & \mathbf 0_3 & \mathbf 0_{3\times2} & \mathbf 0_{3\times2} \\
\mathbf 0_3 & \mathbf 0_3 & \mathbf 0_{3\times2} & \mathbf 0_{3\times2}  \\ 
\frac{ (\bm p_1 - \bm p_c)\times}{\bm I_G}  & \frac{ (\bm p_2 - \bm p_c)\times}{\bm I_G} & 
\frac{\mathbf L}{\bm I_G}   & \frac{ \mathbf L}{\bm I_G} \\
\frac{\mathbf {I}_{3}}{m_{trunk}}  & \frac{ \mathbf {I}_{3}}{m_{trunk}} &  \mathbf {0}_{3\times2} &  \mathbf {0}_{3\times2}  \end{array}  \right]
\end{align}
where $s$ denotes sine operator, and $c$ denotes cosine operator. Note that $\mathbf R_z$ is simplified by the assumption of small roll and pitch angles $\phi \approx 0, \: \theta \approx 0$. \cite{li2021force}

 In equation (\ref{eq:B}), $\bm {I}_G \in \mathbb{R}^{3\times3}$ represent the rotation inertia of the rigid body in the world frame. $\bm p_n$ represents the Cartesian coordinate of the contact point on $n$th foot. $\mathbf L$ is the selection matrix to enforce the 5-D control input, $\mathbf L = [0, 0; 1, 0; 0, 1]$.

\subsection{Optimization Problem Formulation}
The adaptive-frequency trajectory optimization is an offline multiple-shooting discretization method \cite{bulirsch2002introduction} to optimize the robot's CoM trajectory, foot placements, and gait period of each step based on the terrain map. It also maintains the linear velocity close to reference input to generate a smoother walking trajectory.

The optimization variable $\mathbf X \in \mathbb{R}^{39(N+1)}$ includes 
\begin{align}
\label{eq:X}
\mathbf X = [\bm x_N ;\:\: \bm p_{N,1} ;\:\: \bm p_{N,2} ; \:\: \mathbf q_N ;\:\: \bm u_N ;\:\: dt_0\dots dt_{N}]
\end{align}
where $ dt_1\dots dt_{N}$ are discrete sampling times between each two time steps with $N$ total time steps. Subscript $N$ indicates the variable is a column vector of length of $N+1$. For bipedal walking gait, we define the total number of time steps the stance leg spends on the ground and the total number of time steps the swing leg spends in the air to be both 5; hence the one complete two-step  gait period consists of 10 time steps. The MPC prediction horizon is also 10 time steps, which means it predicts a full cycle of periodic gait. It is important to ensure every 5 $dt_i$s has the same length, and thereby the gait period of each step $l$ is the summation of 5 time steps. 

The formulation of the nonlinear programming (NLP) problem is as follows. The optimization objective is to drive the linear velocity close to the command and minimize the ground reaction force to maximize efficiency. 

\begin{align}
\label{eq:cost}
\underset{\bm{x},\:dt_1\dots dt_{N} }{\operatorname{minimize}} \:\: \sum_{i = 0}^{N} \bm \alpha_i(\bm{ \dot p}_{c,x}[i] -\bm  {\dot p}_{c,x}^{ref})^2 + \bm u[i]^\intercal \bm \beta _i \bm u[i] 
\end{align}

\begin{subequations}
\begin{align}
\label{eq:cons1}
\operatorname{s.t.} \:\: \text{Initial Condition}:\:\bm x_0 = \bm x[0] \\
\label{eq:cons2}
\:\: \text{End Condition}:\:\bm x_N = \bm x[N] \\
\label{eq:cons3}
\text{Simplified Dynamics: equation (\ref{eq:simpDyn})} \\
\label{eq:cons4}
\:\: \text{Periodic Gait Constraint} \\
\label{eq:cons5}
\mathbf q_{min} \leq \mathbf q_n[i] = \texttt{IK}(\bm x[i],\: \bm p_n[i]) \leq \mathbf q_{max} \\
\label{eq:cons6}
 \bm \tau_{min} \leq \bm J_n^\intercal(\mathbf q_n[i])\bm u_n[i] \leq \tau_{max} \\
 \label{eq:cons7}
 \bm p_l(\texttt{terrain}) = \bm p_{n,l} = \bm p_{hip,l}+\frac{t_{stance}}{2}\bm { \dot p}_{c,l} \\
  \label{eq:cons8}
 0.02 \leq dt_i \leq 0.05
 \end{align}
\end{subequations}

Equation (\ref{eq:cons4}) enforces the periodic walking gait of the bipedal robot with 5 time steps stance phase and 5 times steps swing phase. Equation (\ref{eq:cons5}) enforces joint angle limits. Equation (\ref{eq:cons6}) constrains joint torques by contact Jacobians. Lastly, the swing foot placement is enforced by the inverted-pendulum-based foot placement policy. (\cite{raibert1986legged,di2018dynamic,li2021force}). With this foot placement policy, the optimization framework can adapt to the most optimal gait period based on how far one step needs to place to overcome the terrain while keeping the robot's linear velocity constant. $t_{stance}$ represents the total time the stance foot spends on the ground, which is the summation of 5 time steps at step $l$. The placement at touch-down for each step $l$ is acclimated to the terrain (i.e. each step is on a stepping stone). 



\section{Adaptive-frequency Control with Varied Gait Periods}
\label{sec:trackingControl}

\begin{figure*}[!t]
	\hspace{0.2cm}
     \center
     \begin{subfigure}[b]{0.78\textwidth}
         \centering
         \includegraphics[width=\textwidth]{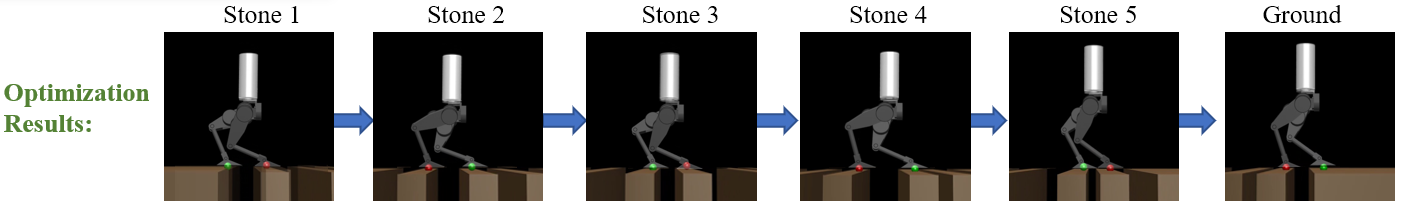}
         \caption{Snapshot of Optimization Results}
         \label{fig:optresults}
     \end{subfigure}
     \\
     \begin{subfigure}[b]{0.85\textwidth}
         \centering
         \includegraphics[clip, trim=0.2cm 0cm 0.3cm 0cm, width=\columnwidth]{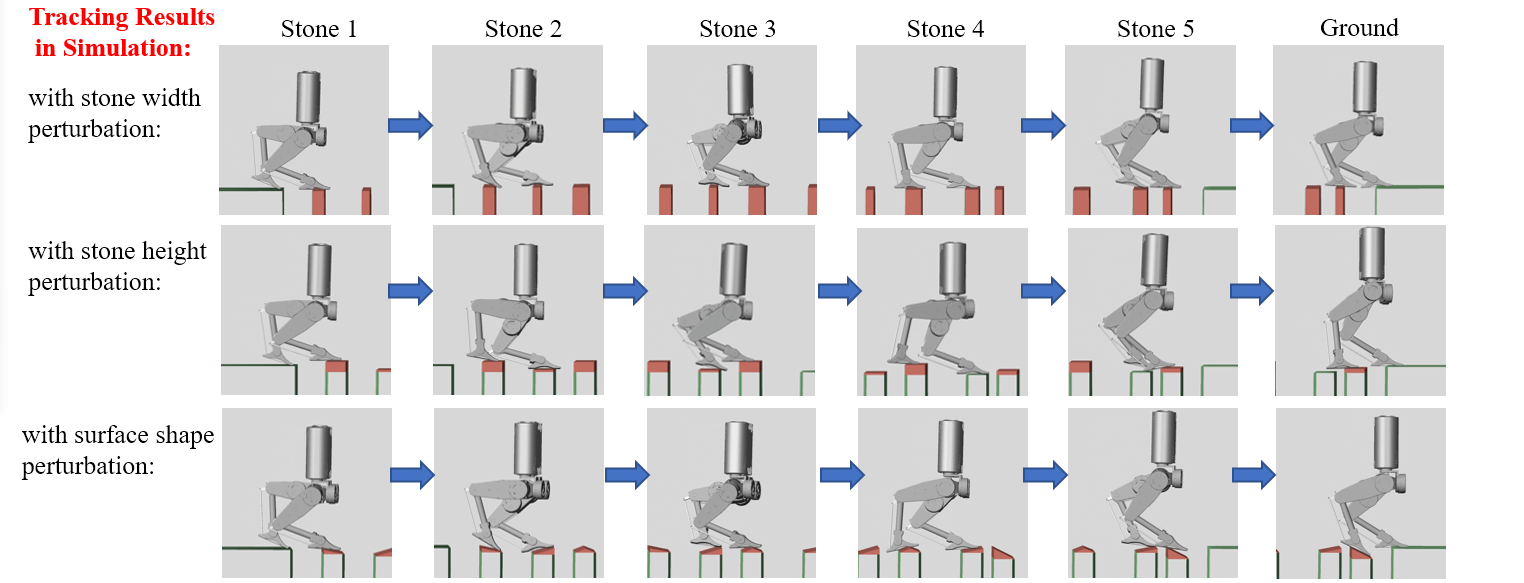}
         \caption{Snapshot of Controller Tracking Results in Simulation with Terrain Perturbations (all using results from (a))}
         \label{fig:trackingresults}
     \end{subfigure}
        \caption{{\bfseries Motion Snapshots of Uneven Stepping Stone Locomotion} a). Optimization results.  b). Simulation results of various cases with terrain perturbations.}
        \label{fig:snapshots}
        \vspace{-1.5em}
\end{figure*}

In this section, we present a force-and-moment-based MPC with adaptive frequency in bipedal walking gait with varied step lengths to overcome discontinued terrains without slowing down or coming to a complete stop.  
 The optimization introduced in Section.\ref{subsec:Optimization} outputs optimized sampling times for MPC, which can also be interpreted as the gait period for each step. Hence it is important to modify these controllers to accept walking gait with different gait periods.

\subsection{Adaptive-frequency MPC for Bipedal Locomotion}
\label{subsec:VGP-MPC}

First, we present the adaptive-frequency MPC. The MPC framework works with varied gait periods from the optimization results. Both MPC and optimization use the same simplified dynamics model shown in Figure. \ref{fig:design}. 

To form a linear state-space dynamics equation for MPC, we choose to include gravity $\bm g$ as a dummy state variable $\bm x = [{\bm \Theta};{\bm p}_c;{\bm \omega};\dot {{\bm p}}_c; \bm g] \in \mathbb{R}^{15}$ in equation (\ref{eq:simpDyn}) to form,  
\begin{align}
\label{eq:linearSS}
\dot{{\bm { x}}}(t) =  {\hat{\bm A_c}} {{\bm {x}}} +  {\hat{\bm B_c}} \bm u.
\end{align}

where continuous-time matrices ${\hat{\bm A_c} \in \mathbb{R}^{15\times15}}$ and ${\hat{\bm B_c} \in \mathbb{R}^{15\times10}}$ are modified from $\bm A$ and $\bm B$.

 A formulation of the MPC problem with finite horizon $k$ can be written in the following form, 
\begin{align}
\label{eq:MPCform}
\underset{\bm{x,u}}{\operatorname{min}}   \:\:  & \sum_{i = 0}^{k-1}(\bm x_{i+1}-  \bm x_{i+1}^{ref})^T\bm Q_i(\bm x_{i+1}- \bm x_{i+1}^{ref}) + \bm{R}_i\| \bm{u}_i \|
\end{align}
\begin{subequations}
\begin{align}
\label{eq:dynamicCons}
\:\:\operatorname{s.t.} \quad  {\bm {x}}[i+1] = \bm {\hat{A}}[i]\bm x[i] + \bm {\hat{B}}[i]\bm u[i], \\
\label{eq:frictionCons}
\nonumber 
-\mu  {F}_{iz} \leq  F_{ix} \leq \mu  {F}_{iz} \quad \quad\\
-\mu  {F}_{iz} \leq  F_{iy} \leq \mu  {F}_{iz} \quad \quad\\
\label{eq:forceCons}
0<  {F}_{min} \leq  F_{iz} \leq  {F}_{max} \quad \quad\\
\label{eq:MPCeqCons}
\bm D_i \bm u_i = 0 \quad \quad \quad \quad
\end{align}
\end{subequations}

The objective of the problem is to drive state $\bm x$ close to command and minimize $\bm u$. These objectives are weighted by diagonal matrices $\bm Q_i\in  \mathbb{R}^{15\times15}$ and $\bm R_i\in \mathbb{R}^{10\times10}$.

Equation (\ref{eq:dynamicCons}) to (\ref{eq:forceCons}) are constraints of the MPC problem. Equation (\ref{eq:dynamicCons}) is an equality constraint of the linearized dynamics equation in discrete-time at $i$th time-step derived from equation (\ref{eq:linearSS}). Equation (\ref{eq:frictionCons}) describes inequality constraints on contact friction pyramid. Equation (\ref{eq:forceCons}) describes the bounds of reaction forces. Equation (\ref{eq:MPCeqCons}) enforces gait constraint to ensure the swing leg exerts zero control input.

The translation of the proposed MPC problem into Quadratic Programming (QP) form to be efficiently solved can be found in many related works and previous works (e.g., \cite{di2018dynamic}, \cite{li2021force}).

\subsection{Whole-body Control}
\label{subsec:WBC}

With adaptive-frequency MPC, in a step with a long gait period, the sampling frequency can be as low as only 20 $\unit{Hz}$. The low-frequency MPC cannot guarantee optimal tracking performance. Hence we choose to combine MPC with WBC to ensure more accurate tracking control. The WBC is an established level-low control method to map reaction forces to joint torques on legged robots \cite{kim2019highly,chignoli2021humanoid}.  

 We adapt the WBC to work with force-and-moment-based MPC control input and allow bipedal walking gait with varied gait periods. The WBCs used in \cite{kim2019highly} and \cite{kim2020dynamic} are paired with a high-frequency joint PD controller to track desired joint position and velocity in addition to computing joint torques based on prioritized tasks. Both CoM and swing foot position control are parts of the WBC tasks. Our WBC framework only uses torque output from QP optimization and does not require joint tracking. Instead, we chose to continue using Cartesian space PD swing foot control \cite{li2021force} to track optimal foot placement from optimization. With this approach, the WBC tasks reduced to only driving CoM position and rotation $\bm x_c = [ p_{c,x},\:  p_{c,y},\:  p_{c,z},\:\phi,\:\theta,\:\psi ]^\intercal$ to desired input (i.e. trajectory tracking). Hence it avoids extra computation time at the very computation-costly derivative of contact Jacobian $\dot{\bm J_c}$ for the 5-DoF bipedal robot leg. 

The full joint space equation of motion for the bipedal robot has the form,
\begin{align}
\label{eq:EOM}
\mathbf M \ddot{\mathbf q} + \mathbf C + \mathbf g = \left[\begin{array}{c}  \mathbf 0 \\  \bm \tau \end{array} \right] 
+ \bm \tau_b
\end{align}
$\ddot{\mathbf q}$ is a linear vector space containing both entries of body state (i.e. CoM position vector and Euler angles) and joint states components, $\ddot{\mathbf q} = [\ddot{\mathbf q}_b;\: \ddot{\mathbf q}_j]$, where $\ddot{\mathbf q}_b \in \mathbb R^6$, $\ddot{\mathbf q}_j \in \mathbb R^{10}$, and $\bm \tau_b = \bm {J}_c^\intercal \bm u$.

The desired acceleration of the CoM tracking task uses the optimal CoM trajectory from the trajectory optimization as reference $\bm x_c^{des}$, and is computed based on a PD control law,
\begin{align}
\label{eq:desAcc}
\ddot{\bm x}_c^{des} = \bm K_P^{WBC}(\bm x_c^{des} - \bm x_c) + \bm K_D^{WBC}(\dot{\bm x}_c^{des} - \dot{\bm x}_c)
\end{align}
And the acceleration command $\ddot {\mathbf {q}}_{cmd}$ is calculated by a similar task-space projection algorithm in \cite{kim2019highly}.

Now the WBC-QP problem to compute the minimized relaxation components of MPC ground reaction force $\Delta \bm u$ and joint acceleration command $\Delta \ddot{\mathbf q}$ is as follows,
\begin{align}
\label{eq:WBC-QP}
\underset{{\Delta \ddot{\mathbf q},\Delta \bm u}}{\operatorname{min}}   \:\:  & 
\Delta \ddot{\mathbf q}^\intercal {\mathbf H} \Delta \ddot{\mathbf q} + \Delta \bm u^\intercal {\mathbf K} \Delta \bm u
\vspace{0.5cm}
\end{align}
\begin{subequations}
\begin{align}
\label{eq:WBC_cons1}
\nonumber
\operatorname{s.t.} \quad 
\mathbf S_{b}\{\mathbf M (\Delta \ddot{\mathbf q} + \ddot{\mathbf q}_{cmd}) + \mathbf C + \mathbf g  \\
- \bm  J_c^\intercal (\Delta \bm  u + \bm u)\} = \mathbf 0 \\
\label{eq:WBC_cons3}
\quad  \quad  \quad  \bm u_{min} \leq \Delta \bm u + \bm u \leq \bm u_{max} \quad\\
\label{eq:WBC_cons4}
\quad  \quad  \quad  \bm \tau_{min} \leq \bm \tau \leq \bm \tau_{max} \quad
\end{align}
\end{subequations}

In equation (\ref{eq:WBC-QP}), $\mathbf H \in \mathbb{R}^{16\times16}$ and $\mathbf K \in  \mathbb{R}^{10\times10}$ are diagonal weighting matrices for each objective. Equation (\ref{eq:WBC_cons1}) is a dynamics constraint to control the floating base dynamics. Selection matrix $\mathbf S_{b}\in  \mathbb{R}^{6\times16}$ consists of 1s and 0s to identify the float base joints. 

The final joint torques can be calculated as
\begin{align}
\label{eq:torque}
\left[\begin{array}{c}  \bm 0 \\  \bm \tau \end{array} \right] 
= \mathbf M (\Delta \ddot{\mathbf q} + \ddot{\mathbf q}_{cmd}) + \mathbf C + \mathbf g - \bm J_c^\intercal (\Delta \bm  u + \bm u)
\end{align}

As for swing leg, the joint torques $\bm {\tau}_{swing,n} \in \mathbb{R}^{5}$ are computed separately by inverse Jacobian $\bm J_{v,n}^\intercal$ of leg $n$,
\begin{align}
\label{eq:forceTorqueMapSwing}
\bm {\tau}_{swing,n} = \bm J_{v,n}^\intercal \bm F_{swing,n}.
\end{align}
Where swing foot force $\bm F_{swing,n}$ is determined by a simple PD control law,
\begin{align}
\label{eq:pdlaw}
\bm F_{swing,n}=\bm K_P(\bm p_{n,des}-\bm p_{n})+\bm K_D(\dot{\bm p}_{n,des}-\dot{\bm p}_n)
\end{align}

\section{Results}
\label{sec:Results}

In this section, we will present highlighted results for validation of our proposed adaptive-frequency control and optimization framework. Associated simulation videos can be found via the link under Figure. \ref{fig:title}.

We validate our proposed approach in a high-fidelity physical-realistic simulation in MATLAB Simulink with Simscape Multibody library. We also use Spatial v2 software package \cite{featherstone2014rigid} to acquire coefficients of dynamics equations in WBC and CasADi \cite{Andersson2019} for offline optimization.  

\begin{figure}[h]
\vspace{0.2cm}
		\center
		\includegraphics[width=1 \columnwidth]{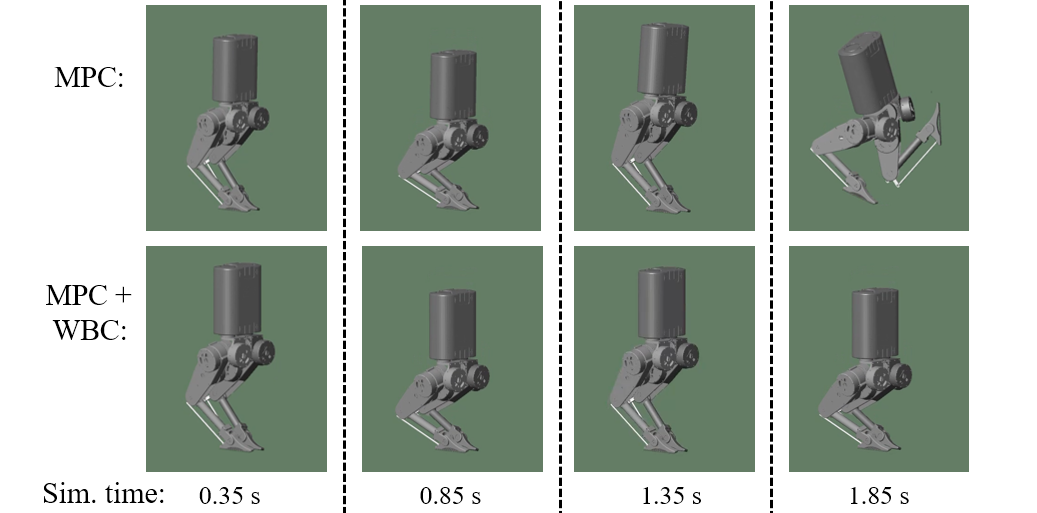}
		\caption{{\bfseries Height Command Tracking Results} Simulation snapshots are several time steps }
		\label{fig:heightcomparison}
		\vspace{-0.2cm}
\end{figure}

Firstly, we present the comparison between MPC-only control vs. MPC+WBC in tracking sinusoidal height command with double-leg stance. Due to low sampling frequency, previous works usually only use MPC as locomotion control and use QP-based force control as balance/stance control for its higher frequency (e.g. \cite{nguyen2019optimized,chignoli2021humanoid,li2021force}). Figure. \ref{fig:heightcomparison} shows the comparison of simulation snapshots between the two approaches, it can be observed that the WBC+MPC approach we proposed performed ideally in height tracking while the MPC-only approach failed over time. 


\begin{figure}[t]
	\vspace{0.2cm}
     \center
     \begin{subfigure}[b]{0.48\textwidth}
         \centering
         \includegraphics[clip, trim=0cm 8.2cm 0cm 0cm, width=\columnwidth]{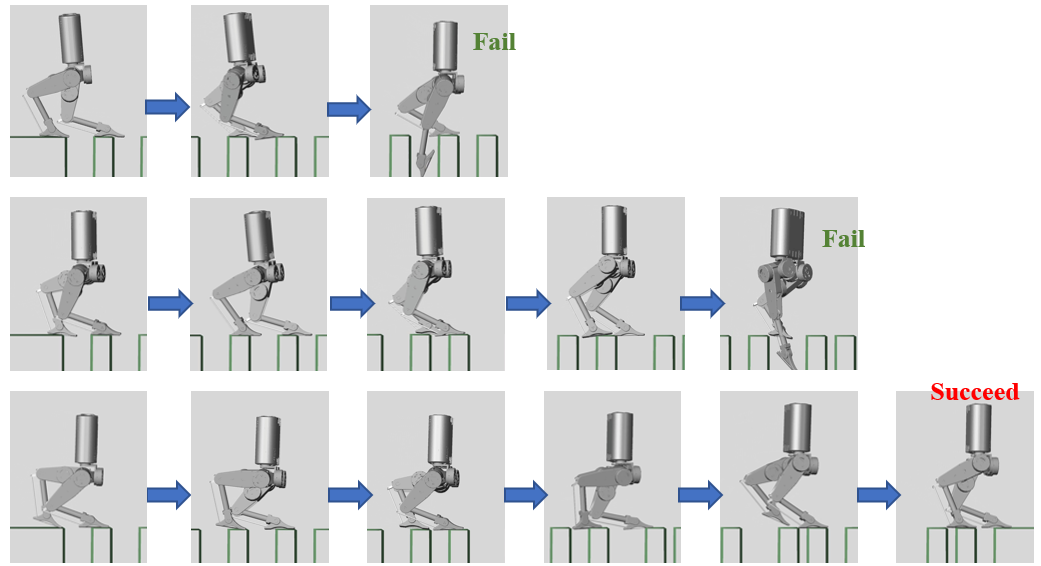}
         \caption{Simulation results: fixed-frequency control}
         \label{fig:fixed_gait_mpc}
     \end{subfigure}
     \\
     \begin{subfigure}[b]{0.48\textwidth}
         \centering
         \includegraphics[clip, trim=0cm 4.1cm 0cm 4.1cm, width=\columnwidth]{Figures/comparison1.png}
         \caption{Simulation results: adaptive-frequency control only}
         \label{fig:no_opt}
     \end{subfigure}
     \\
     \begin{subfigure}[b]{0.48\textwidth}
         \centering
         \includegraphics[clip, trim=0cm 0cm 0cm 8.1cm, width=\columnwidth]{Figures/comparison1.png}
         \caption{Simulation results: adaptive-frequency control + optimization (proposed approach)}
         \label{fig:with_opt}
     \end{subfigure}
        \caption{{\bfseries Motion Snapshots of Uneven Stepping Stone Locomotion} Comparison of fixed-frequency control vs. adaptive-frequency control vs. adaptive-frequency control + optimization }
        \label{fig:snapshots}
        \vspace{-0.0cm}
\end{figure}

\begin{figure}[!h]
\vspace{0.2cm}
		\center
		\includegraphics[width=1 \columnwidth]{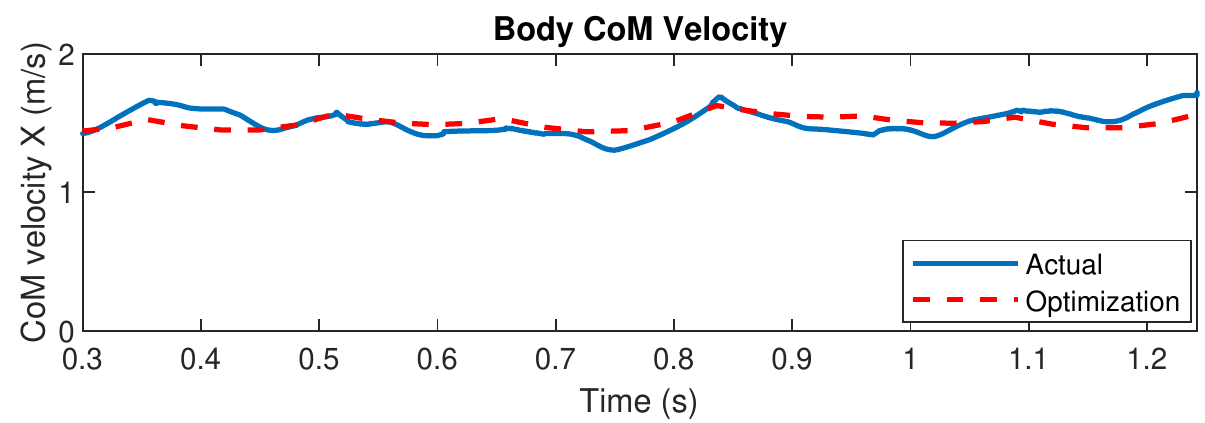}
		\caption{{\bfseries Velocity Tracking Results} Simulation with perturbed stone shapes }
		\label{fig:velocity_tracking}
\end{figure}

Secondly, we compare the locomotion performance over stepping stones in simulation with the following approaches. 
\begin{enumerate}
    \item With fixed-frequency MPC + WBC, at 0.3s
    \item With adaptive-frequency MPC + WBC
    \item With adaptive-frequency MPC + WBC + optimization
\end{enumerate}
As can be seen in Figure. \ref{fig:fixed_gait_mpc}, the approach with fixed-frequency control cannot adapt the foot placement based on the stepping stone gap distance. In Figure. \ref{fig:no_opt}, the adaptive-frequency MPC+WBC framework with manually-input gait periods based on the terrain shows improvement from the fixed gait period case. However, it cannot achieve precise foot placement on stepping stones nor maintain a preferable trajectory, therefore failed after only a few stones. Our proposed approach, shown in Figure. \ref{fig:with_opt}, with both adaptive-frequency control and optimization can allow the bipedal robot to traverse through the stepping stone terrain. Figure. \ref{fig:velocity_tracking} shows the velocity tracking performance with our proposed approach 3). The simulation velocity stays smoothly close to the desired trajectory through the stepping stone terrain.

We also would like to present the solver computation times for several tasks in CasADi with IPOPT solver in MATLAB R2021b. As a benchmark, the PC platform we use for offline optimization has an AMD Ryzen 5-5600X CPU clocked at 4.65$\unit {GHz}$. In Table.\ref{tab:solvingTime}, we measure the solving time of the proposed adaptive-frequency trajectory optimization. The cases are categorized into the number of stepping stones in the terrain. We run the optimization with 30 randomized terrain setups for each case and compute the average time. 

Lastly, we present the uneven stepping stone terrain locomotion results with our proposed approach. 
In realistic scenarios, the stepping stone surface shapes, heights, and widths may vary, hence the errors and disturbances in a vision-based terrain map acquisition system may hinder the accuracy of terrain information. In our approach, we can allow the terrain map in the optimization framework to be simplified to uniformly sized stepping stones with varied center-to-center distances, shown in Figure. \ref{fig:optresults}. We then use this optimization result to control the robot to traverse the terrains with various perturbations, shown in Figure. \ref{fig:trackingresults}. These terrain perturbations including varied stepping stone widths, heights, and surface shapes.
In the above simulation results, the linear velocity the robot maintained during the task is 1.5 $\unit{m/s}$. The stone center-to-center gap distance is between 15 $\unit{cm}$ to 30 $\unit{cm}$.The maximum stone height perturbation is 5 $\unit{cm}$. The stone width perturbation varied between 4 $\unit{cm}$ to 10 $\unit{cm}$.

\begin{table}[!t]
	\vspace{0.2cm}
	\centering
	\caption{Offline Optimization Solving Time}
	\label{tab:solvingTime}
	\begin{tabular}{ccccc}
		\hline
		 Cases: & 4 stones & 5 stones & 6 stones & 7 stones\\
		\hline
		Solving time: & 6.72$\unit{s}$ & 7.93$\unit{s}$ &  10.15$\unit{s}$ & 12.23$\unit{s}$ \\
		\hline 
	\end{tabular}
	\vspace{-0.3cm}
\end{table}	






\section{Conclusions}
\label{sec:Conclusion}

In conclusion, we introduced an effective adaptive-frequency MPC and optimization framework for bipedal locomotion over terrains with discontinuities such as stepping stones with varied gait periods and step lengths. In addition, we also introduced the adaptive-frequency trajectory optimization framework to generate optimal gait periods for each step, CoM trajectory, and foot positions based on the terrain. We paired MPC with WBC for more accurate tracking control performance. Through numerical validation in simulation, we successfully allowed the robot to walk over a series of uneven stepping stones with perturbations while maintaining the robot's average linear velocity at 1.5 $\unit{m/s}$.

\balance
\bibliographystyle{ieeetr}
\bibliography{reference.bib}

\end{document}